\documentclass[runningheads]{llncs}
\usepackage{graphicx}
\begin{document}
\title{Industrial Surface Defect Detection via Diffusion Generation and Asymmetric Student-Teacher Network
	\thanks{Shuo Feng and Runlin Zhou contributed equally to this work.}
	\thanks{\textsuperscript{*} Corresponding author.}
}

\author{
	Shuo Feng\inst{1}
	\and
	Runlin Zhou\inst{1}
	\and
	Yuyang Li\inst{1}
	\and
	Guangcan Liu\inst{1}\textsuperscript{*}
}

\authorrunning{S. Feng et al.}

\institute{
	School of Automation, Southeast University, Nanjing, China \\
	\email{220252099@seu.edu.cn}
}
\maketitle
\begin{abstract}
Industrial surface defect detection often suffers from limited defect samples, severe long-tailed distributions, and difficulties in accurately localizing subtle defects under complex backgrounds. To address these challenges, this paper proposes an unsupervised defect detection method that integrates a Denoising Diffusion Probabilistic Model (DDPM) with an asymmetric teacher-student architecture.
First, at the data level, the DDPM is trained solely on normal samples. By introducing constant-variance Gaussian perturbations and Perlin noise-based masks, high-fidelity and physically consistent defect samples along with pixel-level annotations are generated, effectively alleviating the data scarcity problem.
Second, at the model level, an asymmetric dual-stream network is constructed. The teacher network provides stable representations of normal features, while the student network reconstructs normal patterns and amplifies discrepancies between normal and anomalous regions.
Finally, a joint optimization strategy combining cosine similarity loss and pixel-wise segmentation supervision is adopted to achieve precise localization of subtle defects.
Experimental results on the MVTecAD dataset show that the proposed method achieves 98.4\% image-level AUROC and 98.3\% pixel-level AUROC, significantly outperforming existing unsupervised and mainstream deep learning methods. The proposed approach does not require large amounts of real defect samples and enables accurate and robust industrial defect detection and localization.

\end{abstract}
\vspace*{-30pt}
\section{Introduction}
\vspace*{-10pt}
With the rapid advancement of intelligent manufacturing and Industry 4.0, industrial quality control has become a critical component in modern production systems. Surface defect detection, as a key stage in quality inspection, directly affects product qualification rates, brand reputation, and economic benefits. In industries such as metal processing, electronics manufacturing, textiles, and glass or plastic forming, surface defects such as scratches, stains, holes, and deformations are inevitable. These defects not only affect product appearance but also compromise structural integrity and service life, potentially leading to safety risks.

Traditional inspection methods mainly rely on manual visual inspection, which suffers from several limitations. First, it is inefficient and cannot keep up with high-speed automated production lines. Second, it is subjective and prone to inconsistencies among inspectors. Third, prolonged visual work leads to fatigue, making it difficult to detect subtle or low-contrast defects. Finally, manual inspection incurs high labor costs and lacks traceability.
In recent years, computer vision and deep learning techniques have become dominant solutions due to their non-contact, high-speed, and high-accuracy advantages. However, despite their success, these methods still face a critical challenge: data scarcity.
In real industrial scenarios, normal samples are abundant, while defect samples are rare and follow a long-tailed distribution. Rare but critical defects are particularly difficult to collect. Moreover, pixel-level annotation of defects is expensive and time-consuming, requiring expert knowledge.This makes the definition of defects not always easy to define at the image level\cite{bhatt2021image}.
Traditional data augmentation methods\cite{niu2020defect}\cite{duan2023few}, such as rotation, flipping, and brightness adjustment, only introduce low-level variations and cannot generate new defect patterns. Furthermore, they may distort intrinsic texture structures, leading to unrealistic samples and poor generalization.
Existing unsupervised defect detection methods, especially reconstruction-based approaches, often fail to detect subtle defects because reconstruction errors can be overwhelmed by background noise. Similarly, conventional teacher-student frameworks usually adopt symmetric architectures without explicit anomaly amplification mechanisms, limiting their localization accuracy.

To address these issues\cite{nichol2021improved}\cite{livernoche2024diffusion}\cite{rombach2022latent}\cite{song2020ddim}\cite{song2020sde}, this paper proposes a unified framework that combines DDPM-based defect generation with an asymmetric teacher-student architecture. The method forms a closed-loop system of “defect generation → high-precision detection.” The DDPM is trained using only normal samples to learn the distribution of normal textures and illumination. During reverse diffusion, constant-variance perturbations and Perlin noise masks are introduced to generate realistic defect regions along with pixel-level masks. Meanwhile, an asymmetric teacher-student network is designed, where the frozen teacher provides a stable normal feature baseline, and the student reconstructs normal features from anomalous inputs, amplifying feature discrepancies. Finally, cosine similarity loss and segmentation supervision are used to achieve precise defect localization.
The main contributions of this paper are summarized as follows:
\vspace*{-10pt}
\begin{enumerate} 
\item A DDPM-based defect generation strategy requiring only normal samples is proposed, enabling high-fidelity and diverse defect synthesis with pixel-level masks.

\item An asymmetric teacher-student network is designed to explicitly amplify anomaly features through reconstruction. Extensive experiments demonstrate superior performance over existing methods on industrial datasets.
\end{enumerate}
\vspace*{-14.5pt}
\section{Related Work}
\vspace*{-12pt}
\subsection{Industrial Surface Defect Detection}
\vspace*{-10pt}
Industrial surface defect detection is a critical component of quality control in intelligent manufacturing. Existing methods can be broadly categorized into three main paradigms:
Traditional machine vision methods rely on handcrafted features (e.g., edge detection\cite{canny1986computational}, texture analysis	\cite{ojala2002multiresolution}). While effective in simple scenarios, they suffer from poor generalization and struggle to adapt to complex industrial environments.
Supervised deep learning methods\cite{ronneberger2015unet}\cite{ren2015faster}, typically based on CNNs, learn discriminative features automatically and achieve high detection accuracy. However, they require large amounts of labeled defect samples and fail to detect previously unseen defect types.
Unsupervised/semi-supervised anomaly detection methods\cite{roth2021patchcore}\cite{wang2021student} are trained exclusively on normal samples and identify anomalies by modeling the distribution of defect-free data. This paradigm aligns perfectly with the scarcity of defective samples in real-world industrial scenarios and has become the mainstream research direction in recent years.
\vspace*{-15pt}
\subsection{Defect Synthesis and Data Augmentation}
\vspace*{-10pt}
Data scarcity is a fundamental bottleneck in industrial defect detection. To address this issue, several representative solutions have been proposed:
Traditional data augmentation	\cite{cubuk2019autoaugment}\cite{cubuk2020randaugment} expands the dataset via geometric and photometric transformations. While easy to implement, it cannot generate new defect patterns.
GAN/VAE-based synthesis\cite{zhang2021defect}\cite{kingma2013auto} leverages generative models to learn the distribution of real defects and produce realistic samples. However, these methods often suffer from training instability and mode collapse.
Pseudo-anomaly synthesis, such as CutPaste\cite{li2021cutpaste} and DRAEM\cite{zavrtanik2021draem}, generates synthetic defects by applying simple transformations to normal images. They are easy to train and have shown promising performance.
Diffusion model-based generation\cite{ho2020denoising}\cite{zhang2023diffusionad} achieves high-quality and diverse defect synthesis, making it the most recent research trend, though at the cost of high computational overhead.
\vspace*{-15pt}
\subsection{Reconstruction-Based and Teacher-Student Anomaly Detection}
\vspace*{-10pt}
Reconstruction-based and teacher-student frameworks are the two most successful paradigms for unsupervised anomaly detection.
Reconstruction-based methods assume that normal samples can be accurately reconstructed, while anomalies lead to large reconstruction errors. Representative approaches include AE/VAE, DRAEM, and DeSTSeg\cite{zhang2023destseg}. Among them, DRAEM introduces pseudo-anomaly samples during training, and DeSTSeg further improves localization accuracy by incorporating a segmentation decoder.
Teacher-student-based methods use a pre-trained large model (e.g., ResNet\cite{wang2021student}, ViT\cite{zhang2023destseg}, CLIP	\cite{jeong2023winclip}) as the teacher network to extract features, while a student network is trained to mimic the teacher’s outputs. Anomalies are then detected based on the discrepancy between the two outputs. This paradigm is easy to train, has fast inference speed, and is among the state-of-the-art approaches currently available.
\vspace*{-25pt}
\subsection{Limitations of Existing Methods}
\vspace*{-8pt}
Despite their success, existing methods still suffer from several critical limitations:
\vspace*{-10pt}
\begin{enumerate}
\item Reconstruction-based methods often struggle with complex textures, leading to high false positive rates.
\item Traditional teacher-student architectures typically adopt a symmetric design, resulting in insufficient knowledge utilization and high computational cost.
\item Diffusion models are mostly used as data augmentation tools, and their generative capabilities are not deeply integrated into the anomaly detection pipeline.
\item Most methods perform well in controlled laboratory environments but fail to adapt to the complex conditions of real-world industrial scenarios.
\end{enumerate}
\vspace*{-10pt}
To address these issues, this paper proposes an industrial surface defect detection method based on DDPM\cite{ho2020denoising}\cite{dhariwal2021diffusion}\cite{rombach2022latent} and an asymmetric teacher-student architecture. By combining the generative power of diffusion models with the discriminative advantages of the teacher-student framework, our method achieves more accurate and efficient defect detection.
\vspace*{-14.5pt}
\section{Method}
\vspace*{-11pt}
\subsection{DDPM-Based Defect Generation}
\vspace*{-8pt}
The generation process consists of three stages:
\vspace*{-4pt}
\begin{enumerate}
\item	Forward diffusion: gradually adds Gaussian noise to images.
\item	Reverse denoising: learns to recover normal images.
\item	Defect synthesis: introduces perturbations to generate anomalies.
\end{enumerate}
\vspace*{-15pt}
\subsubsection{Forward Diffusion}
The forward diffusion process is a noise-adding process. Let the original real defect-free or defective image distribution be \(x_0 \sim q(x)\). The forward diffusion is a fixed-parameter Markov chain that gradually injects Gaussian noise into the image over \(T\) discrete time steps according to a predefined variance schedule \(\beta_1, \beta_2, \dots, \beta_T\). For a given previous state \(x_{t-1}\), the transition probability of the current state \(x_t\) is defined as:
\setlength{\abovedisplayskip}{-2pt}
\setlength{\belowdisplayskip}{-20pt}   
\begin{equation}
q(x_t \mid x_{t-1}) = \mathcal{N}\left(x_t ; \sqrt{1-\beta_t}x_{t-1},\ \beta_t I\right)
\end{equation}
\setlength{\abovedisplayskip}{10pt}
\setlength{\belowdisplayskip}{10pt}
\vspace*{-10pt}
\subsubsection{Reverse Denoising Process}
Data generation is essentially the inverse solution of the noise-adding process. Since directly computing the true posterior probability \(q(x_{t-1} \mid x_t)\) is intractable in high-dimensional space, DDPM uses a deep neural network \(p_\theta\) parameterized by \(\theta\) to approximate this reverse Markov chain. When \(\beta_t\) is sufficiently small, the reverse transition distribution also follows a Gaussian distribution, and its mathematical model can be expressed as:
\setlength{\abovedisplayskip}{-2pt}
\setlength{\belowdisplayskip}{-18pt}   
\begin{equation}
p_\theta(x_{t-1} \mid x_t) = \mathcal{N}\left(x_{t-1}; \mu_\theta(x_t, t),\ \Sigma_\theta(x_t, t)\right)
\end{equation}
\vspace*{-10pt}
\subsubsection{Defect Generation Process}
In the reverse denoising stage, a Perlin noise generator is used to capture diverse random anomaly topologies with smooth edges, which are binarized into a local defect mask matrix \(M\). Subsequently, an opacity parameter \(\delta\) is introduced to perform pixel-level weighted blending between the original real normal image \(I\) and the generated global anomaly image \(P\). The final synthetic local defect image is expressed as Fig.1:
\setlength{\abovedisplayskip}{0pt}
\setlength{\belowdisplayskip}{0pt}
\begin{equation}
A = \overline{M} \odot I + (1-\delta)(M \odot I) + \delta(M \odot P)
\end{equation}
\setlength{\abovedisplayskip}{10pt}
\setlength{\belowdisplayskip}{10pt}
where $\overline{M} = 1 - M$ denotes the inverted mask, and $\odot$ represents the element-wise multiplication operation.
\begin{figure}[ht]
	\includegraphics[width=\textwidth]{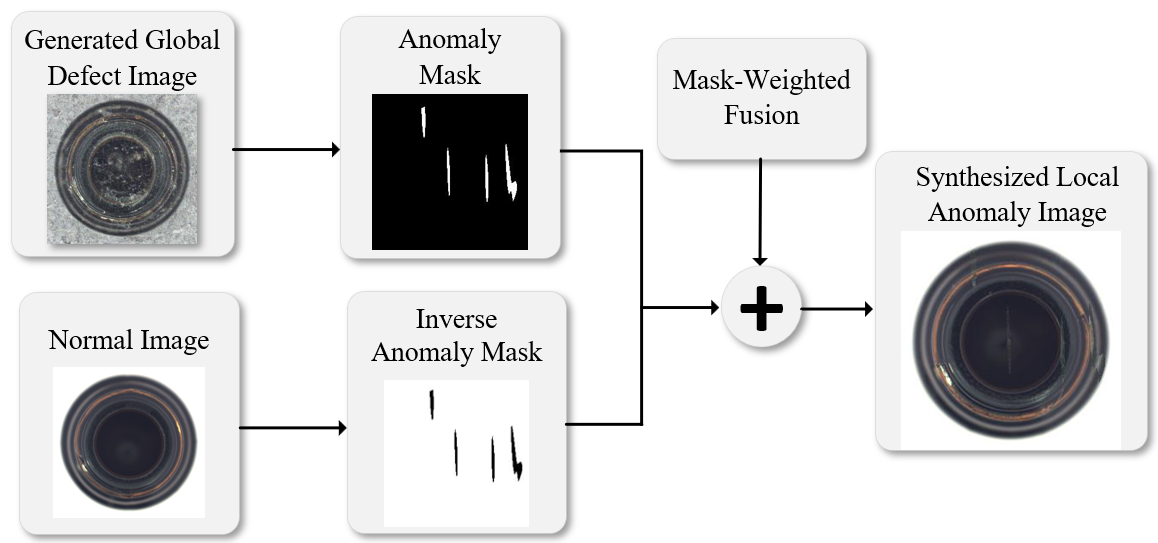}
	\caption{The overall pipeline of the local defect synthesis method based on Perlin noise mask.} \label{fig1}
\end{figure}
\subsection{Dataset Reconstruction Scheme for Downstream Denoising and Defect Detection Tasks}
\vspace*{-10pt}

To better adapt to downstream denoising and defect detection tasks, this paper proposes a structured triplet-based sample construction strategy. During the data organization and loading stage, the indexing mechanism of raw samples is redesigned to spatially bind the normal image $I$, the synthesized local defect image $A$, and the corresponding binary defect mask $M$ generated by the previous synthesis pipeline. As a result, a stable triplet representation $(I, A, M)$ is constructed.

For mask processing, the original continuous-valued defect masks are binarized into a clear foreground-background partition (0-1 representation). This operation enhances the clarity of the supervisory signal, improves the discriminative ability in pixel-level learning tasks, and promotes more stable convergence during model training.

Furthermore, in some implementations, morphological operations such as erosion and dilation can be applied to refine the boundaries of the defect masks. This further improves boundary quality and structural consistency of the annotated regions, thereby enhancing the model's sensitivity to fine-grained defect structures.
\vspace*{-15pt}
\subsection{Asymmetric Teacher--Student Architecture for Defect Detection}
\vspace*{-6pt}
\paragraph {3.3.1 Overall Network Architecture}
This section proposes an end-to-end asymmetric teacher-student dual-stream architecture based on the Vision Transformer (ViT). The architecture decouples normal feature extraction
and anomalous feature reconstruction into two independent branches through a structurally
asymmetric design.

\textbf{Teacher Network:}  
The teacher network adopts a frozen, pre-trained ViT encoder, which extracts stable and unbiased normal representations. It serves as a reference feature extractor and is not updated during training.

\textbf{Student Network:}  
The student network consists of a ViT encoder and a dedicated deep decoder in an asymmetric structure. It is responsible for reconstructing defect images by mapping them back to normal texture distributions, thereby producing feature discrepancies when compared with the teacher network.

\textbf{Dual-Stream Feature Alignment:}  
Multi-level features from the teacher and student networks are strictly aligned. Channel-wise $L_2$ normalization is applied to eliminate magnitude effects, ensuring that only directional differences in feature space are preserved. This provides a stable basis for anomaly scoring.

\textbf{Segmentation Head:}  
The segmentation head converts high-dimensional feature discrepancies into pixel-level anomaly probability maps. By learning the spatial distribution of defects, it outputs pixel-wise confidence scores indicating abnormality, enabling precise defect localization and segmentation.

\vspace*{-10pt}
\paragraph{3.3.2 Multi-Task Jointly Driven Denoising Training Strategy}
To fully exploit the anomaly localization and denoising capabilities of the asymmetric dual-stream architecture, a multi-task jointly driven training strategy is proposed. This strategy aims to guide the student network to learn a stable mapping from anomalous representations to the normal manifold in the latent space. Meanwhile, pixel-level mask supervision is incorporated to enable fine-grained localization and segmentation of local defects.
\begin{figure}[ht]
	\includegraphics[width=\textwidth]{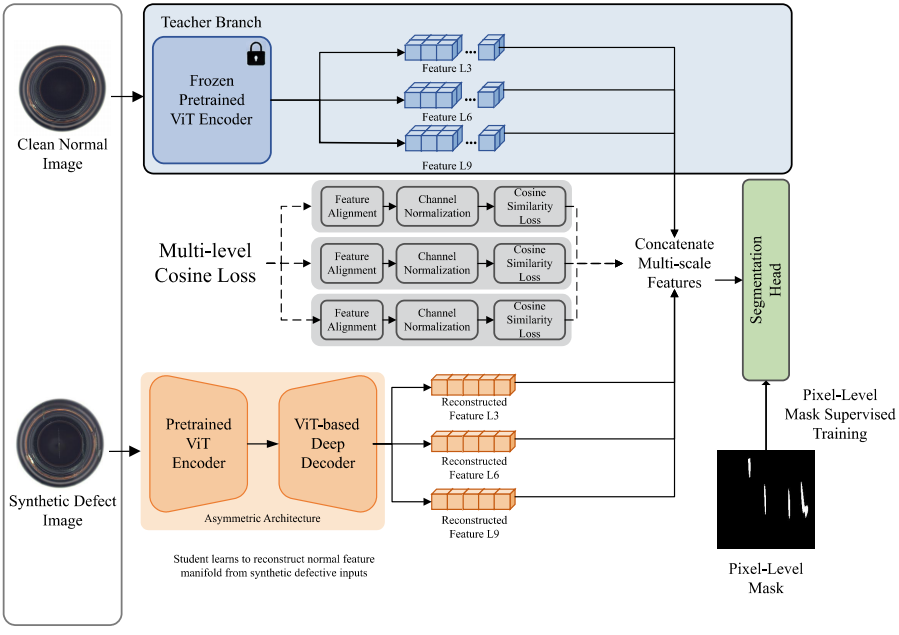}
	\caption{Multi-Task Jointly Driven Training Process.} \label{fig2}
\end{figure}
\begin{figure}[ht]
	\includegraphics[width=\textwidth]{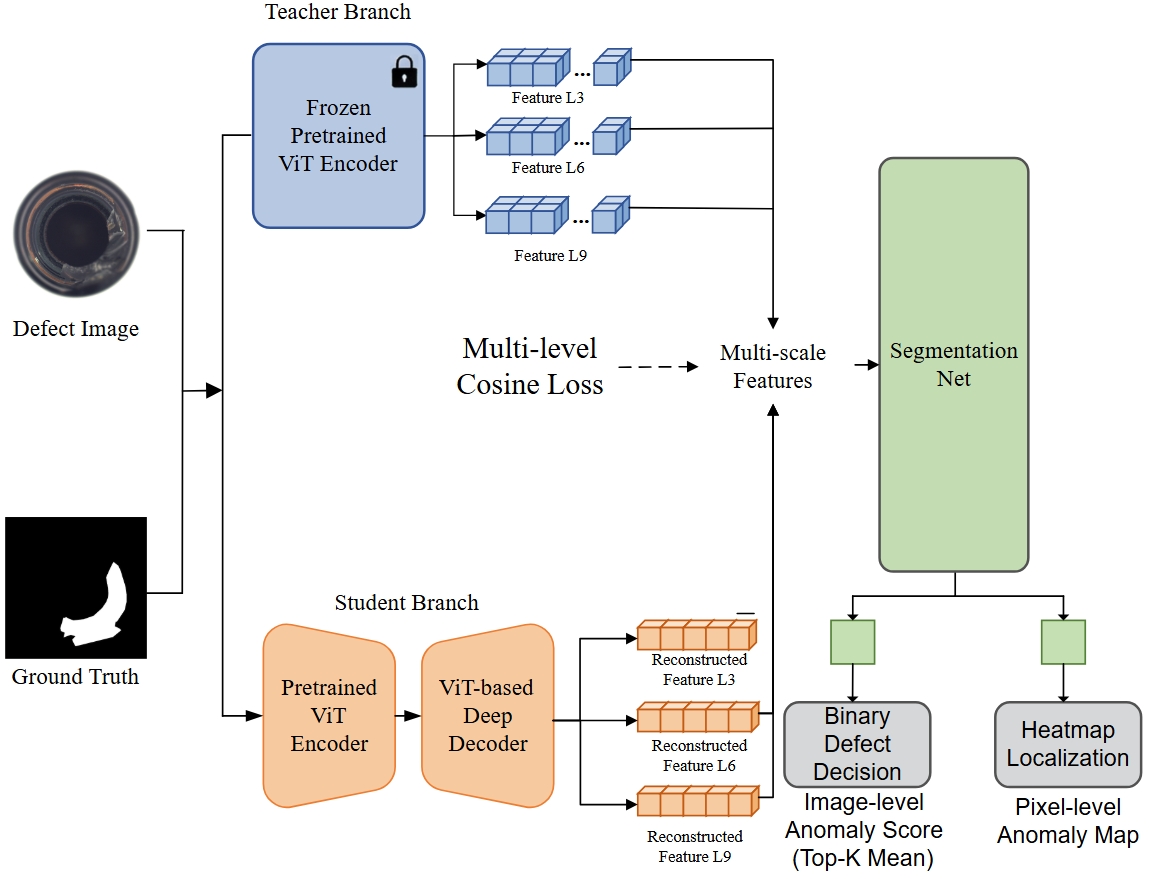}
	\caption{Multi-Task Jointly Driven Inference Process.} \label{fig3}
\end{figure}
\paragraph{3.3.3 Loss Function}
The core training objective of our asymmetric dual-stream framework is to enforce the student decoder to produce feature representations that closely match those extracted by the frozen teacher network on clean defect-free images. Leveraging cross-layer spatial topology alignment and channel-wise $L_2$ normalization, the spatial distance between features is rigorously converted into a measure of semantic direction discrepancy in the high-dimensional feature space.
Specifically, for the l-th key depth layer selected in the network architecture (e.g., shallow, middle, and deep layers),
let $\hat{F}_T^l \in R^{C \times H \times W}$ denote the $L_2$-normalized reference feature tensor of the teacher network,
and $\hat{F}_S^l \in R^{C \times H \times W}$ denote the corresponding decoded reconstruction feature tensor of the student network.
To penalize the semantic direction deviation between the two in the feature space, this study introduces a pixel-wise cosine similarity loss.
At any spatial coordinate $(h, w)$,Since the scales of the feature vectors have been unified, their cosine distance can be directly obtained by computing the dot product of the two vectors and subtracting this dot product from 1. Therefore, the mathematical definition of the feature reconstruction loss $L_{cos}^l$ for the $l$-th layer is as follows:
\setlength{\abovedisplayskip}{-2pt}
\setlength{\belowdisplayskip}{-2pt}
\begin{equation}
	L_{cos}^l = \frac{1}{H \times W} \sum_{h=1}^{H} \sum_{w=1}^{W} \left( 1 - \sum_{c=1}^{C} \hat{F}_{T,c}^l(h, w) \cdot \hat{F}_{S,c}^l(h, w) \right)
\end{equation}
\setlength{\abovedisplayskip}{10pt}
\setlength{\belowdisplayskip}{10pt}
\vspace*{-15pt}
\subsection{Defect Localization}
\vspace*{-8pt}
After training the asymmetric teacher-student network, the inference stage achieves a complete pipeline from image input to final defect localization through three steps: feature difference calculation, anomaly map generation, and anomaly score aggregation. This section details the anomaly score calculation method and the spatial localization inference mechanism.
\vspace*{-10pt}
\paragraph{3.4.1 Feature Difference Map Generation in Testing Phase}
During testing, the input image is fed into both the teacher and student networks simultaneously. The teacher network extracts raw visual features, while the student network performs denoising and normal texture reconstruction for defective regions. After channel-wise $L_2$ normalization, the pixel-wise feature difference between the teacher and student is calculated to obtain the anomaly difference map at each layer:
\setlength{\abovedisplayskip}{-5pt}
\setlength{\belowdisplayskip}{-5pt}
\begin{equation}
	A^l(h,w) = 1 - \sum_{c=1}^{C} \hat{F}_{T,c}^l(h,w) \cdot \hat{F}_{S,c}^l(h,w)
\end{equation}
\setlength{\abovedisplayskip}{10pt}
\setlength{\belowdisplayskip}{10pt}
\vspace*{-2pt}
The difference approaches 0 for normal regions and increases significantly for defective regions, forming a clear anomaly response matrix.
\vspace*{-10pt}
\paragraph{3.4.2 Pixel-Level Anomaly Score Extraction}
After multi-scale fusion and output by the segmentation head, a probability matrix $M$ within the range $[0, 1]$ is obtained. This property avoids the cumbersome extreme value normalization in traditional methods. Each value in the output matrix can be directly defined as the pixel-level anomaly score at the corresponding spatial position, which can be directly used for pixel-level localization and AUROC calculation.
\vspace*{-10pt}
\paragraph{3.4.3 Image-Level Anomaly Score Aggregation}
To improve robustness in industrial scenarios, Top-K mean aggregation is adopted to obtain the image-level anomaly score, suppressing isolated noise and false responses. After sorting the pixel scores in descending order, the top K maximum values are averaged:
\setlength{\abovedisplayskip}{-2pt}
\setlength{\belowdisplayskip}{-2pt}
\begin{equation}
	S_I = \frac{1}{K} \sum_{i=1}^{K} s_i
\end{equation}

In this work, $K=100$. This method effectively highlights real continuous defects and completes the final anomaly detection and spatial localization.
\vspace*{-10pt}
\section{Experiments}
\vspace*{-8pt}
\subsection{Datasets and Experimental Setup}
\vspace*{-8pt}
This paper adopts MVTec AD,VisA,MPDD as the core benchmark dataset for evaluation. MVTec AD consists of 5 texture categories and 10 object categories from industrial manufacturing, with a total of 5354 images. The dataset is composed of normal images for training, and both normal and anomalous images with various defects for testing. It also provides pixel-level annotations for defective test images.
All experiments in this paper are conducted under the following computational environment. The hardware platform includes an Intel Xeon Platinum 8358P CPU, an NVIDIA A800 80GB GPU, 1TB of RAM, and 39TB of storage space. The software environment is based on the Ubuntu 22.04.3 LTS operating system and CUDA 12.1. Model training and testing are implemented using Python 3.11.0 and the PyTorch deep learning network.
\vspace*{-20pt}
\subsection{Evaluation Metrics}
\vspace*{-10pt}
In the industrial defect detection task, this paper adopts three mainstream unsupervised evaluation metrics to measure the model performance from three perspectives: image-level detection, pixel-level localization, and region-wise balance:

\textbf{Image-level AUROC (I-AUROC)}\cite{bergmann2019mvtec}:
This metric evaluates the model's binary classification ability to determine whether a defect exists in the entire image. A higher value indicates a higher accuracy of image-level anomaly detection.

\textbf{Pixel-level AUROC (P-AUROC)}\cite{bergmann2020uninformed}:
This metric measures the pixel-level segmentation accuracy of the model for defective regions, reflecting the model's localization ability for defect positions and boundaries.

\textbf{Area Under Per-Region Overlap (AUPRO)}\cite{schneider2020pro}:
This metric provides a balanced evaluation of the localization ability for defects of different sizes, avoiding the dominance of large defects in the metric, which is more aligned with the detection requirements of subtle defects in industrial scenarios.
\vspace*{-15pt}
\subsection{Anomaly Detection Results on MVTec AD}
\vspace*{-10pt}
The anomaly detection results on the MVTec AD dataset are summarized in Table 1,Table 2,Table 3. The anomaly detection results on the Visa,MPDD dataset are summarized in Table 4,Table 5. 

\begin{table}[htbp]
	\centering
	\caption{Per-category I-AUROC results on MVTec AD dataset.}
	\label{tab:auroc}
	\renewcommand{\arraystretch}{0.8}
	\setlength{\tabcolsep}{0pt}
	{\scriptsize
	\begin{tabular}{lcccccccc}
		\hline	
		Category & \tiny SimpleNet\cite{liu2023simplenet} & \tiny RealNet\cite{zhang2024realnet} & \tiny CFLOW-AD\cite{gudovskiy2022cflow} & \tiny PyramidalAD\cite{lei2023pyramidflow} & \tiny DiAD\cite{he2023diad} & \tiny UniAD\cite{chen2022utrad} &\tiny DeSTSeg\cite{zhang2023destseg} & Ours \\
		\hline
		Bottle & 100.0 & 95.6 & 99.9 & 77.1 & 97.7 & 99.4 &100.0& \textbf{100.0}  \\
		Cable & 97.1 & 70.0 & 89.6 & 58.2 & 88.5 & 93.0 & 97.5 & \textbf{99.0}  \\
		Capsule & 85.4 & 64.4 & 86.1 & 55.6 & 88.6 & 69.8 & 90.3 & \textbf{96.8}  \\
		Hazelnut & 100.0 &100.0 & 99.9 & 92.1 & 96.9 & 99.6 & 99.6 & \textbf{100.0}   \\
		MetalNut & 99.0 & 78.6 & 96.9 & 63.0 & 88.1 & 98.0 & 99.8 & \textbf{100.0}   \\
		Carpet & 97.1 & 96.5 & 99.2 & 44.6 & 64.5 & 99.8 & 95.9 & \textbf{99.7}  \\
		Grid & 98.2 & 97.2 & 86.7 & 75.5 & 99.0 & 99.0 & 99.6 & \textbf{100.0}   \\
		Leather & 100.0 & 100.0 & 100.0 & 67.2 & 97.6 & 100.0 & 100.0 & \textbf{100.0}  \\
		Tile & 100.0 & 97.5 & 99.8 & 78.6 & 93.4 & 96.7 & 99.7 & \textbf{100.0}  \\
		Wood & 99.5 & 99.6 & 98.4 & 86.7 & 94.3 & 98.2 & 98.6 & \textbf{99.2}  \\
		Pill & 90.6 & 68.5 & 84.3 & 67.0 & 94.7 & 75.1 & 92.7 & \textbf{98.4}   \\
		Screw & 77.5 & 71.3 & 63.4 & 63.7 & 63.2 & \textbf{88.4} & 86.1 & 86.0  \\
		Toothbrush & 89.4 & 76.4 & 84.4 & 89.4 & 95.3 & 83.9 & 91.7 & \textbf{100.0} \\
		Transistor & 97.8 & 79.7 & 90.5 & 50.8 & \textbf{99.0} & 99.3 & 96.9 & 96.5  \\
		Zipper & 99.1 & 77.3 & 95.6 & 83.3 & 73.5 & 86.9 & 97.6 & \textbf{100.0} \\
		\hline
		\textbf{Average} & 95.4 & 84.8 & 91.6 & 70.2 & 88.9 & 92.5 & 96.4 & \textbf{98.4} \\
		\hline
	\end{tabular}
}
\end{table}
\begin{table}[htbp]
	\centering
	\caption{Per-category P-AUROC results on MVTec AD dataset.}
	\label{tab:aupro}
	\scriptsize
	\renewcommand{\arraystretch}{0.75}
	\setlength{\tabcolsep}{2pt}
	\begin{tabular}{lcccccccc}
		\hline
		Category & SimpleNet & RealNet & CFLOW-AD & PyramidalAD & DiAD & UniAD & DeSTSeg & Ours \\
		\hline
		Bottle & 97.4 & 69.8 & 97.4 & 77.7 & 93.5 & 98.1 & 94.9 & \textbf{99.5}  \\
		Cable & 96.6 & 61.5 & 89.8 & 82.8 & 90.7 & 96.1 & 95.2 & \textbf{97.6}  \\
		Capsule & 98.1 & 54.6 & 98.5 & 90.3 & 90.9 & 97.5 & 89.3 & \textbf{98.9}  \\
		Hazelnut & 98.1 & 77.5 & 98.5 & 92.7 & 95.3 & 98.0 & 96.8 & \textbf{99.6}  \\
		MetalNut & 97.7 & 52.5 & 96.0 & 81.6 & 94.5 & 93.4 & 95.8 & \textbf{98.7} \\
		Carpet & 97.7 & 89.2 & 98.8 & 79.2 & 87.9 & 98.5 & 95.7 & \textbf{96.5}  \\
		Grid & 96.9 & 82.6 & 92.9 & 85.7 & 85.7 & 94.8 & 96.9 & \textbf{99.4}  \\
		Leather & 98.5 & 97.9 & 99.2 & 87.7 & 90.6 & 99.1 & 99.3 & \textbf{99.8}  \\
		Tile & 95.4 & 93.9 & 96.0 & 75.7 & 76.1 & 89.2 & 97.5 & \textbf{99.4}  \\
		Wood & 92.5 & 90.4 & 94.2 & 62.6 & 82.8 & 93.1 & 93.2 & \textbf{97.0}  \\
		Pill & 96.7 & 54.4 & 96.7 & 83.3 & 92.8 & 90.0 & 90.1 & \textbf{99.3}  \\
		Screw & 95.8 & 51.8 & 96.5 & 71.4 & 86.3 & \textbf{98.2} & 76.2 & 95.3 \\
		Toothbrush & 98.0 & 84.8 & 98.2 & 73.0 & 89.6 & 98.4 & 95.6 & \textbf{99.5} \\
		Transistor & 95.4 & 60.9 & 84.8 & 75.9 & \textbf{97.8} & 97.3 & 73.9 & 94.9  \\
		Zipper & 97.9 & 67.6 & 97.9 & 81.0 & 84.4 & 95.8 & 90.6 & \textbf{98.6} \\
		\hline
		\textbf{Average} & 96.8 & 72.6 & 95.7 & 80.0 & 89.3 & 95.8 & 92.0 & \textbf{98.3} \\
		\hline
	\end{tabular}
\end{table}
\begin{table}[htbp]
	\centering
	\caption{Per-category AUPRO results on MVTec AD dataset.}
	\label{tab:f1}
	\scriptsize
	\renewcommand{\arraystretch}{0.75}
	\setlength{\tabcolsep}{2pt}
	\begin{tabular}{lcccccccc}
		\hline
		Category & SimpleNet & RealNet & CFLOW-AD & PyramidalAD & DiAD & UniAD & DeSTSeg & Ours \\
		\hline
		Bottle & 89.1 & 60.9 & 91.9 & 40.9 & 76.0 & 95.3 & 91.8 & \textbf{98.2} \\
		Cable & 86.0 & 33.3 & 78.0 & 41.6 & 63.9 & 84.9 & 85.1 & \textbf{99.0} \\
		Capsule & 87.1 & 23.4 & 92.8 & 57.3 & 54.5 & 86.9 & 60.6 & \textbf{96.5} \\
		Hazelnut & 93.9 & 75.4 & 95.6 & 84.2 & 81.3 & 93.7 & 92.9 & \textbf{97.1} \\
		MetalNut & 87.6 & 39.6 & 88.5 & 37.3 & 53.5 & 80.8 & 94.8 & \textbf{97.8} \\
		Carpet & 90.0 & 84.0 & 94.3 & 52.3 & 67.0 & \textbf{95.9} & 93.4 & 93.2 \\
		Grid & 88.3 & 77.7 & 81.0 & 66.9 & 56.7 & 89.3 & 91.9 & \textbf{97.5}\\
		Leather & 95.5 & 98.0 & 98.1 & 74.0 & 68.3 & 98.3 & 98.4 & \textbf{99.5} \\
		Tile & 82.5 & 90.5 & 86.5 & 34.4 & 49.7 & 77.6 & 95.7 & \textbf{98.6} \\
		Wood & 80.0 & 88.8 & 90.2 & 32.3 & 57.5 & 89.5 & 95.0 & \textbf{95.8} \\
		Pill & 85.3 & 35.1 & 90.5 & 65.2 & 65.2 & 89.2 & 70.2 & \textbf{98.8} \\
		Screw & 83.1 & 18.5 & 87.7 & 21.5 & 57.4 & \textbf{90.9} & 53.3 & 84.9\\
		Toothbrush & 80.6 & 34.1 & 84.5 & 23.2 & 65.3 & 85.8 & 65.5 & \textbf{97.0} \\
		Transistor & 82.5 & 44.6 & 73.0 & 26.1 & 85.7 & \textbf{92.2} & 76.6 & 84.2 \\
		Zipper & 91.9 & 47.7 & 92.4 & 55.7 & 56.7 & 88.7 & 86.2 & \textbf{95.8} \\
		\hline
		\textbf{Average} & 86.9 & 56.8 & 88.3 & 47.5 & 63.9 & 89.3 & 83.4 & \textbf{95.6} \\
		\hline
	\end{tabular}
\end{table}
\begin{table}[htbp]
	\centering
	\caption{Performance comparison with state-of-the-art models on the VisA dataset.}
	\label{tab:visa}
	\begin{tabular}{lccc}
		\hline
		Model & I-AUROC & P-AUROC & AUPRO \\
		\hline
		SimpleNet & 86.4 & 96.6 & 79.2 \\
		RealNet & 71.4 & 61.0 & 27.4 \\
		CFLOW-AD & 86.5 & 96.7 & \underline{86.8} \\
		PyramidalFlow & 58.2 & 77.0 & 42.8 \\
		DiAD & 84.8 & 82.5 & 44.5 \\
		UniAD & 89.0 & \underline{97.3} & 86.5 \\
		DeSTSeg & \underline{89.9} & 86.7 & 61.1 \\
		Ours & \textbf{92.3} & \textbf{97.4} & \textbf{89.0} \\
		\hline
	\end{tabular}
\end{table}
\begin{table}[htbp]
	\centering
	\caption{Performance comparison with state-of-the-art models on the MPDD dataset.}
	\label{tab:mpdd}
	\begin{tabular}{lccc}
		\hline
		Method & I-AUROC & P-AUROC & AUPRO \\
		\hline
		SimpleNet & 88.4 & 95.5 & 89.0 \\
		RealNet & 85.1 & 83.3 & 68.1 \\
		CFLOW-AD & 75.7 & 95.8 & \underline{89.5} \\
		PyramidalFlow & 73.6 & 94.1 & 77.2 \\
		DiAD & 68.3 & 90.4 & 66.1 \\
		UniAD & 70.5 & 93.9 & 79.7 \\
		DeSTSeg & \underline{91.3} & 82.0 & 63.3 \\
		Ours & \textbf{95.8} & \underline{95.6} & \textbf{91.9} \\
		\hline
	\end{tabular}
\end{table}
\vspace*{-15pt}
Compared with mainstream methods such as SimpleNet, CFLOW‑AD and DeSTSeg, the proposed model achieves significant improvements in localization accuracy for weak defects, slender scratches and small-target defects, yielding superior overall performance.
\vspace*{-15pt}
\subsection{Anomaly Localization on MVTec AD}
\vspace*{-10pt}
In the anomaly localization task on the MVTec AD dataset, this paper adopts pixel-level AUROC (P-AUROC) and the region-balanced metric AUPRO for quantitative evaluation. Across all categories of the MVTec AD dataset, the proposed method achieves a pixel-level AUROC of 98.3\% and an AUPRO of 95.6\%. It outperforms the comparison methods in the localization of subtle defects, slender defects, and small-target defects, demonstrating stable and high-precision spatial localization capabilities for industrial defects. Representative samples of anomaly localization are visualized, as shown in Figure 4 .
\vspace*{-10pt}
\begin{figure}[ht]
	\includegraphics[width=\textwidth]{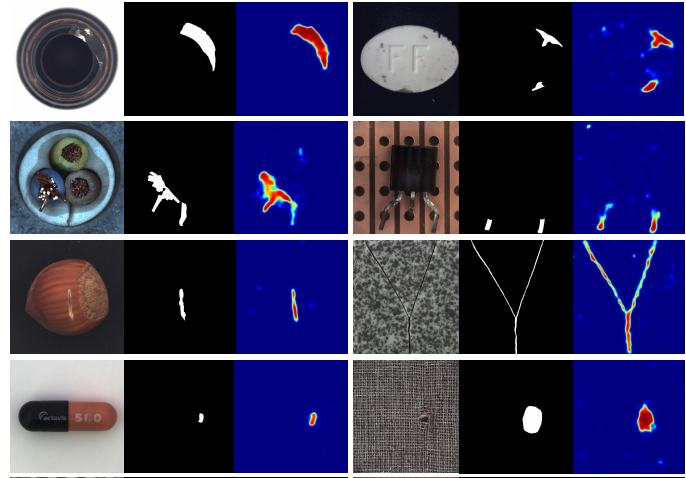}
	\caption{Visualization of defect localization results on the MVTec AD dataset.} \label{fig4}
\end{figure}
\vspace*{-30pt}
\subsection{Ablation Study}
\vspace*{-7pt}
To verify the influence of each core module on detection performance,
this chapter conducts an ablation study based on the MVTec AD dataset.
Key components including the decoder, cosine loss, segmentation head,
and loss function are removed or replaced respectively to quantify
the necessity of each structure.

Experimental results show that the decoder, cosine similarity constraint,
segmentation head, and L1 loss are all vital parts of the network.
Missing any module will lead to a decline in localization accuracy.
Experimental results are shown in Table 4.
\begin{table}[ht]
	\centering
	\caption{Ablation study on core network modules.}
	\label{tab:ablation_modules}
	\small
	\setlength{\tabcolsep}{4pt}
	\begin{tabular}{lccccccc}
		\hline
		Config & Decoder & Cosine & Seg & Focal & I-AUROC & P-AUROC & PRO \\
		\hline
		w/o Decoder      &  -  &  Y  &  Y  &  Y  & 96.3 & 97.5 & 94.2 \\
		w/o Cosine Loss  &  Y  &  -  &  Y  &  Y  & 97.5 & 97.6 & 94.1 \\
		w/o Seg Head     &  Y  &  Y  &  -  &  -  & 98.1 & 97.9 & 92.4 \\
		w/o Focal Loss   &  Y  &  Y  &  Y  &  -  & 98.2 & 97.4 & 94.5 \\
		Ours (Full)      &  Y  &  Y  &  Y  &  Y  & \textbf{98.4} & \textbf{98.3} & \textbf{95.6} \\
		\hline
	\end{tabular}
\end{table}
\vspace*{-8pt}
\paragraph{Decoder}
The decoder provides denoising and reconstruction capabilities for the student network, which is the key to generating feature differences. After removing the decoder, the student network cannot complete the normalization reconstruction of abnormal regions, leading to a significant decline in the overall performance of the model, with the AUPRO dropping to 94.2\%.
\vspace*{-8pt}
\paragraph{Cosine Similarity Loss (Cosine Loss)}
The cosine loss is used to align the features of the teacher and student networks and resist light and texture disturbances. After removing this constraint, the model is easily interfered by the background, and the image-level detection index (I-AUROC) drops to 97.5\%.
\vspace*{-10pt}
\paragraph{Segmentation Head}
The segmentation head is responsible for converting high-dimensional differences into pixel-level anomaly scores. After removing the segmentation head, the model cannot output sharp defect boundaries, and its fine-grained localization ability is significantly reduced, with the AUPRO dropping to 92.4\%.
\vspace*{-10pt}
\paragraph{L1 Loss}
The use of L1 loss can make the segmentation boundaries sharper and more sensitive to small defects. When replaced with a common loss, the edge segmentation accuracy decreases, and the AUPRO drops by about 4.3 percentage points.

In summary, the asymmetric dual-stream structure, denoising decoder, cosine feature alignment, and segmentation head supervision cooperate with each other to jointly achieve high-precision and high-robustness industrial defect detection and pixel-level localization.

\vspace*{-15pt}
\section{Conclusion}
\vspace*{-10pt}
This paper conducts a joint study on data generation and defect detection to address the challenges in industrial surface defect detection, such as the scarcity of real abnormal samples, long-tailed distribution of defects, and difficulty in localizing subtle defects.
First, a high-fidelity defect generation scheme relying only on normal samples is implemented based on the Denoising Diffusion Probabilistic Model (DDPM), which effectively alleviates the problem of insufficient defect samples.
Then, an asymmetric teacher-student architecture for defect denoising and detection is proposed. Through feature denoising reconstruction and cross-level feature alignment, the representation discrepancy between normal and abnormal regions is amplified.
Combined with segmentation head supervision, high-precision defect detection and pixel-level spatial localization are achieved.

Experimental results on the standard MVTec AD dataset show that the proposed method achieves excellent performance in image-level detection, pixel-level localization, and region-balanced metrics.
It can effectively identify common industrial defects including subtle scratches, tiny defects, and low-contrast stains, with strong anti-interference ability and generalization.
Ablation studies verify the key roles of the decoder, cosine similarity constraint, segmentation head, and loss function in improving detection accuracy and localization performance.

The proposed scheme achieves a good balance among detection accuracy, model generalization, and engineering practicality.
It provides a feasible solution for industrial surface defect detection with few-shot conditions, complex textures, and high-precision requirements, and also offers effective technical support for practical production line deployment.

\section*{Acknowledgements}
This work was supported in part by
National Natural Science Foundation of China (NSFC) under Grant U21B2027 and the Start-up Research Fund of Southeast University under Grant RF1028623006.

\vspace*{-10pt}

%
%
%

\begin{thebibliography}{99}
	
	\bibitem{bhatt2021image}
	Bhatt P M, Malhan R K, Rajendran P, et al. Image-based surface defect detection using deep learning: A review[J].
	Journal of Computing and Information Science in Engineering, 2021, 21(4): 040801.
	
	\bibitem{niu2020defect}
	Niu S, Wang Y, Wang F, et al. Defect image sample generation with GAN for improving defect recognition[J].
	IEEE Transactions on Automation Science and Engineering, 2020, 18(3): 1071--1082.
	
	\bibitem{duan2023few}
	Duan Y, Liu J, Wang Z, et al. Few-shot defect image generation via defect-aware feature manipulation[C].
	Proceedings of the AAAI Conference on Artificial Intelligence, 2023: 1346--1354.
	
	\bibitem{nichol2021improved}
	Nichol A Q, Dhariwal P. Improved denoising diffusion probabilistic models[C].
	International Conference on Machine Learning, 2021: 8162--8171.
	
	\bibitem{livernoche2024diffusion}
	Livernoche V, Köhler L, Eisenbacher M, et al. On diffusion modeling for anomaly detection[C].
	Proceedings of the IEEE/CVF Winter Conference on Applications of Computer Vision, 2024: 2032--2041.
	
	\bibitem{rombach2022latent}
	Rombach R, Blattmann A, Lorenz R, et al. High-resolution image synthesis with latent diffusion models[C].
	Proceedings of the IEEE/CVF Conference on Computer Vision and Pattern Recognition, 2022: 10684--10695.
	
	\bibitem{song2020ddim}
	Song J, Meng C, Ermon S. Denoising diffusion implicit models[C].
	International Conference on Learning Representations, 2020.
	
	\bibitem{song2020sde}
	Song Y, Sohl-Dickstein J, Kingma D P, et al. Score-based generative modeling through stochastic differential equations[C].
	International Conference on Learning Representations, 2020.
	
	\bibitem{ho2020denoising}
	Ho J, Jain A, Abbeel P. Denoising diffusion probabilistic models[C].
	Advances in Neural Information Processing Systems, 2020, 33: 6840--6851.
	
	\bibitem{dhariwal2021diffusion}
	Dhariwal P, Nichol A. Diffusion models beat GANs on image synthesis[C].
	Advances in Neural Information Processing Systems, 2021, 34: 8780--8794.
	
	\bibitem{canny1986computational}
	Canny J. A computational approach to edge detection[J].
	IEEE Transactions on Pattern Analysis and Machine Intelligence, 1986, 8(6): 679--698.
	
	\bibitem{ojala2002multiresolution}
	Ojala T, Pietikainen M, Maenpaa T. Multiresolution gray-scale and rotation invariant texture classification with local binary patterns[J].
	IEEE Transactions on Pattern Analysis and Machine Intelligence, 2002, 24(7): 971--987.
	
	\bibitem{ronneberger2015unet}
	Ronneberger O, Fischer P, Brox T. U-net: Convolutional networks for biomedical image segmentation[C].
	International Conference on Medical Image Computing and Computer-Assisted Intervention, 2015: 234--241.
	
	\bibitem{ren2015faster}
	Ren S, He K, Girshick R, et al. Faster r-cnn: Towards real-time object detection with region proposal networks[C].
	Advances in Neural Information Processing Systems, 2015, 28: 91--99.
	
	\bibitem{roth2021patchcore}
	Roth K, Pemula L, Zepeda J, et al. Patchcore: Towards total recall in industrial anomaly detection[C].
	Proceedings of the IEEE/CVF International Conference on Computer Vision, 2021: 14009--14019.
	
	\bibitem{wang2021student}
	Wang G, Han S, Ding E, et al. Student-teacher feature pyramid matching for anomaly detection[C].
	Proceedings of the British Machine Vision Conference, 2021: 1--14.
	
	\bibitem{cubuk2019autoaugment}
	Cubuk E D, Zoph B, Mane D, et al. Autoaugment: Learning augmentation policies from data[C].
	Proceedings of the IEEE/CVF Conference on Computer Vision and Pattern Recognition, 2019: 113--123.
	
	\bibitem{cubuk2020randaugment}
	Cubuk E D, Zoph B, Shlens J, et al. Randaugment: Practical automated data augmentation with a reduced search space[C].
	Proceedings of the IEEE/CVF Conference on Computer Vision and Pattern Recognition Workshops, 2020: 702--703.
	
	\bibitem{zhang2021defect}
	Zhang G, Cui K, Hung T Y, et al. Defect-gan: High-fidelity defect synthesis for automated defect inspection[C].
	Proceedings of the IEEE/CVF Winter Conference on Applications of Computer Vision, 2021: 2515--2524.
	
	\bibitem{kingma2013auto}
	Kingma D P, Welling M. Auto-encoding variational bayes[J].
	arXiv preprint arXiv:1312.6114, 2013.
	
	\bibitem{li2021cutpaste}
	Li C L, Sohn K, Yoon J, et al. Cutpaste: Self-supervised learning for anomaly detection and localization[C].
	Proceedings of the IEEE/CVF Conference on Computer Vision and Pattern Recognition, 2021: 9664--9674.
	
	\bibitem{zavrtanik2021draem}
	Zavrtanik V, Kristan M, Skocaj D. Draem-a discriminatively trained reconstruction embedding for surface anomaly detection[C].
	Proceedings of the IEEE/CVF International Conference on Computer Vision, 2021: 8330--8339.
	
	\bibitem{zhang2023diffusionad}
	Zhang H, Wang Z, Wu Z, et al. Diffusionad: Denoising diffusion for anomaly detection[C].
	Proceedings of the IEEE/CVF Conference on Computer Vision and Pattern Recognition, 2023: 843--852.
	
	\bibitem{zhang2023destseg}
	Zhang X, Li S, Li X, et al. Destseg: Segmentation guided denoising student-teacher for anomaly detection[C].
	Proceedings of the IEEE/CVF Conference on Computer Vision and Pattern Recognition, 2023: 3914--3923.
	
	\bibitem{jeong2023winclip}
	Jeong J, Kim S, Seo D, et al. Winclip: Zero-/few-shot anomaly classification and segmentation[C].
	Proceedings of the IEEE/CVF Conference on Computer Vision and Pattern Recognition, 2023: 19613--19622.
	
	\bibitem{bergmann2019mvtec}
	Bergmann P, Fauser M, Sattlegger D, et al. MVTec AD -- A comprehensive real-world dataset for unsupervised anomaly detection[C].
	Proceedings of the IEEE/CVF Conference on Computer Vision and Pattern Recognition, 2019: 9592--9600.
	
	\bibitem{bergmann2020uninformed}
	Bergmann P, Fauser M, Sattlegger D, et al. Uninformed students: Student-teacher anomaly detection with discriminative latent features[C].
	Proceedings of the IEEE/CVF Conference on Computer Vision and Pattern Recognition, 2020: 8771--8780.
	
	\bibitem{schneider2020pro}
	Schneider T, Bergmann P, Steger C. The per-region overlap (PRO) score: A fair evaluation metric for anomaly localization[J].
	arXiv preprint arXiv:2009.14067, 2020.
	
	\bibitem{liu2023simplenet}
	Liu Z, Wang Y, Han Y, et al. SimpleNet: A simple network for image anomaly detection and localization[C].
	Proceedings of the IEEE/CVF Conference on Computer Vision and Pattern Recognition, 2023: 20402--20411.
	
	\bibitem{zhang2024realnet}
	Zhang X, Xu M, Zhou X. RealNet: A feature selection network with realistic synthetic anomaly for anomaly detection[C].
	Proceedings of the IEEE/CVF Conference on Computer Vision and Pattern Recognition, 2024: 23678--23687.
	
	\bibitem{gudovskiy2022cflow}
	Gudovskiy D, Ishizaka S, Kozuka K. CFLOW-AD: Real-time unsupervised anomaly detection with localization via conditional normalizing flows[C].
	Proceedings of the IEEE/CVF Winter Conference on Applications of Computer Vision, 2022: 1434--1442.
	
	\bibitem{lei2023pyramidflow}
	Lei J, Hu X, Wang Y, et al. PyramidFlow: High-resolution defect contrastive localization using pyramid normalizing flow[C].
	Proceedings of the IEEE/CVF Conference on Computer Vision and Pattern Recognition, 2023: 14143--14152.
	
	\bibitem{he2023diad}
	He H, Zhang J, Chen H, et al. DiAD: A diffusion-based framework for multi-class anomaly detection[J].
	arXiv preprint arXiv:2312.06607, 2023.
	
	\bibitem{chen2022utrad}
	Chen L, You Z, Zhang N, et al. UTRAD: Anomaly detection and localization with U-Transformer[J].
	Neural Networks, 2022, 147: 53--62.
	
\end{thebibliography}
%

\end{document}